\documentclass{article}
\usepackage[accepted]{icml2017}
\usepackage{times}
\usepackage[utf8]{inputenc}
\usepackage[T1]{fontenc}
\usepackage{booktabs}
\usepackage{nicefrac}
\usepackage{microtype}
\usepackage{amsmath}
\usepackage{amsthm}
\usepackage{amssymb}
\usepackage{amsfonts}
\usepackage{graphicx}
\usepackage{subfigure} 
\usepackage{makecell}
\usepackage{url}
\usepackage{hyperref}

\icmltitlerunning{Deep Transfer Learning with Joint Adaptation Networks}

\begin{document}

\twocolumn[
\icmltitle{Deep Transfer Learning with Joint Adaptation Networks}

\begin{icmlauthorlist}
\icmlauthor{Mingsheng Long}{tsinghua}
\icmlauthor{Han Zhu}{tsinghua}
\icmlauthor{Jianmin Wang}{tsinghua}
\icmlauthor{Michael I. Jordan}{berkeley}
\end{icmlauthorlist}

\icmlaffiliation{tsinghua}{Key Lab for Information System Security, MOE; Tsinghua National Lab for Information Science and Technology (TNList); NEL-BDS; School of Software, Tsinghua University, Beijing 100084, China}
\icmlaffiliation{berkeley}{University of California, Berkeley, Berkeley 94720}

\icmlcorrespondingauthor{Mingsheng Long}{mingsheng@tsinghua.edu.cn}

\icmlkeywords{Transfer learning, Deep learning, Kernel embedding, Adversarial learning}

\vskip 0.3in
]

\printAffiliationsAndNotice{}

\begin{abstract} 
Deep networks have been successfully applied to learn transferable features for adapting models from a source domain to a different target domain. In this paper, we present joint adaptation networks (JAN), which learn a transfer network by aligning the joint distributions of multiple domain-specific layers across domains based on a joint maximum mean discrepancy (JMMD) criterion. Adversarial training strategy is adopted to maximize JMMD such that the distributions of the source and target domains are made more distinguishable. Learning can be performed by stochastic gradient descent with the gradients computed by back-propagation in linear-time. Experiments testify that our model yields state of the art results on standard datasets.
\end{abstract} 

\section{Introduction}
Deep networks have significantly improved the state of the arts for diverse machine learning problems and applications. Unfortunately, the impressive performance gains come only when massive amounts of labeled data are available for supervised learning. Since manual labeling of sufficient training data for diverse application domains on-the-fly is often prohibitive, for a target task short of labeled data, there is strong motivation to build effective learners that can leverage rich labeled data from a different source domain. However, this learning paradigm suffers from the shift in data distributions across different domains, which poses a major obstacle in adapting predictive models for the target task \cite{cite:Book09DSS,cite:TKDE10TLSurvey}. 

Learning a discriminative model in the presence of the shift between training and test distributions is known as transfer learning or domain adaptation \cite{cite:TKDE10TLSurvey}. Previous shallow transfer learning methods bridge the source and target domains by learning invariant feature representations or estimating instance importance without using target labels \cite{cite:NIPS06KMM,cite:TNN11TCA,cite:ICML13Landmark}. Recent deep transfer learning methods leverage deep networks to learn more transferable representations by embedding domain adaptation in the pipeline of deep learning, which can simultaneously disentangle the explanatory factors of variations behind data and match the marginal distributions across domains \cite{cite:Arxiv14DDC,cite:ICCV15SDT,cite:ICML15DAN,cite:NIPS16RTN,cite:ICML15RevGrad,cite:NIPS16DSN}.

Transfer learning becomes more challenging when domains may change by the joint distributions of input features and output labels, which is a common scenario in practical applications. First, deep networks generally learn the complex function from input features to output labels via multilayer feature transformation and abstraction. Second, deep features in standard CNNs eventually transition from general to specific along the network, and the transferability of features and classifiers decreases when the cross-domain discrepancy increases \cite{cite:NIPS14CNN}. Consequently, after feed-forwarding the source and target domain data through deep networks for multilayer feature abstraction, the shifts in the joint distributions of input features and output labels still linger in the network activations of multiple domain-specific higher layers. Thus we can use the joint distributions of the activations in these domain-specific layers to approximately reason about the original joint distributions, which should be matched across domains to enable domain adaptation. To date, this problem has not been addressed in deep networks.

In this paper, we present Joint Adaptation Networks (JAN) to align the joint distributions of multiple domain-specific layers across domains for unsupervised domain adaptation. JAN largely extends the ability of deep adaptation networks \cite{cite:ICML15DAN} to reason about the joint distributions as mentioned above, while keeping the training procedure even simpler. Specifically, JAN admits a simple transfer pipeline, which processes the source and target domain data by convolutional neural networks (CNN) and then aligns the joint distributions of activations in multiple task-specific layers. To learn parameters and enable alignment, we derive joint maximum mean discrepancy (JMMD), which measures the Hilbert-Schmidt norm between kernel mean embedding of empirical joint distributions of source and target data. Thanks to a linear-time unbiased estimate of JMMD, we can easily draw a mini-batch of samples to estimate the JMMD criterion, and implement it efficiently via back-propagation. We further maximize JMMD using adversarial training strategy such that the distributions of source and target domains are made more distinguishable. Empirical study shows that our models yield state of the art results on standard datasets.

\section{Related Work}
Transfer learning \cite{cite:TKDE10TLSurvey} aims to build learning machines that generalize across different domains following different probability distributions \cite{cite:NIPS08CSA,cite:TNN11TCA,cite:TPAMI12DTMKL,cite:ICML13Landmark,cite:ICML13TCS}. Transfer learning finds wide applications in computer vision \cite{cite:ECCV10Office,cite:ICCV11GFS,cite:CVPR12GFK,cite:NIPS14LSDA} and natural language processing \cite{cite:JMLR11MTLNLP,cite:ICML11DADL}. 

The main technical problem of transfer learning is how to reduce the shifts in data distributions across domains. Most existing methods learn a shallow representation model by which domain discrepancy is minimized, which cannot suppress domain-specific exploratory factors of variations. Deep networks learn abstract representations that disentangle the explanatory factors of variations behind data \cite{cite:TPAMI13DLSurvey} and extract transferable factors underlying different populations \cite{cite:ICML11DADL,cite:CVPR13MidLevel}, which can only reduce, but not remove, the cross-domain discrepancy \cite{cite:NIPS14CNN}. Recent work on deep domain adaptation embeds domain-adaptation modules into deep networks to boost transfer performance \cite{cite:Arxiv14DDC,cite:ICCV15SDT,cite:arXiv17ADDA,cite:ICML15RevGrad,cite:ICML15DAN,cite:NIPS16RTN}. These methods mainly correct the shifts in marginal distributions, assuming conditional distributions remain unchanged after the marginal distribution adaptation.

Transfer learning will become more challenging as domains may change by the joint distributions $P({\bf X},{\bf Y})$ of input features $\mathbf{X}$ and output labels $\mathbf{Y}$. The distribution shifts may stem from the marginal distributions $P({\bf X})$ (a.k.a. covariate shift \cite{cite:NIPS06KMM,cite:NIPS08CSA}), the conditional distributions $P({\bf Y}|{\bf X})$ (a.k.a. conditional shift \cite{cite:ICML13TCS}), or both (a.k.a. dataset shift \cite{cite:Book09DSS}). Another line of work \cite{cite:ICML13TCS,cite:NIPS14FTL} correct both target and conditional shifts based on the theory of kernel embedding of conditional distributions \cite{cite:ICML09HSE,cite:ICML10HSE,cite:JMLR10HSE}. Since the target labels are unavailable, adaptation is performed by minimizing the discrepancy between marginal distributions instead of conditional distributions. In general, the presence of conditional shift leads to an ill-posed problem, and an additional assumption that the conditional distribution may only change under location-scale transformations on ${\bf X}$ is commonly imposed to make the problem tractable \cite{cite:ICML13TCS}. As it is not easy to justify which components of the joint distribution are changing in practice, our work is transparent to diverse scenarios by directly manipulating the joint distribution without assumptions on the marginal and conditional distributions. Furthermore, it remains unclear how to account for the shift in joint distributions within the regime of deep architectures.

\section{Preliminary}

\subsection{Hilbert Space Embedding}
We begin by providing an overview of Hilbert space embeddings of distributions, where each distribution is represented by an element in a reproducing kernel Hilbert space (RKHS). Denote by $\mathbf{X}$ a random variable with domain $\Omega$ and distribution $P(\mathbf{X})$, and by $\mathbf{x}$ the instantiations of $\mathbf{X}$.
A reproducing kernel Hilbert space (RKHS) $\mathcal{H}$ on $\Omega$ endowed by a kernel ${k\left( {{\mathbf{x}},{\mathbf{x'}}} \right)}$ is a Hilbert space of functions $f : \Omega \mapsto \mathbb{R}$ with inner product ${\left\langle { \cdot , \cdot } \right\rangle _\mathcal{H}}$. Its element ${k\left( {{\mathbf{x}}, \cdot } \right)}$ satisfies the reproducing property: ${\left\langle {f\left(  \cdot  \right),k\left( {{\mathbf{x}}, \cdot } \right)} \right\rangle _\mathcal{H}} = f\left( {\mathbf{x}} \right)$. Alternatively, ${k\left( {{\mathbf{x}}, \cdot } \right)}$ can be viewed as an (infinite-dimensional) implicit feature map ${\phi \left( {\mathbf{x}} \right)}$ where $k\left( {{\mathbf{x}},{\mathbf{x'}}} \right) = {\left\langle {\phi \left( {\mathbf{x}} \right),\phi \left( {{\mathbf{x'}}} \right)} \right\rangle _\mathcal{H}}$. 
Kernel functions can be defined on vector space, graphs, time series and structured objects to handle diverse applications. The kernel embedding represents a probability distribution $P$ by an element in RKHS endowed by a kernel $k$ \cite{cite:ALT07KSE,cite:JMLR10HSE,cite:JMLR12MMD}
\begin{equation}\label{eqn:Ex}
	{\mu _{\mathbf{X}}}\left( P \right) \triangleq {\mathbb{E}_{\mathbf{X}}}\left[ {\phi \left( {\mathbf{X}} \right)} \right] = \int\nolimits_\Omega  {\phi \left( {\mathbf{x}} \right){\text{d}}P\left( {\mathbf{x}} \right)}, 
\end{equation}
where the distribution is mapped to the expected feature map, i.e. to a point in the RKHS, given that ${\mathbb{E}_{\mathbf{X}}}\left[ {k\left( {{\mathbf{x}},{\mathbf{x'}}} \right)} \right] \leqslant \infty $. The mean embedding ${\mu _{\mathbf{X}}}$ has the property that the expectation of any RKHS function $f$ can be evaluated as an inner product in $\mathcal{H}$, ${\left\langle {{\mu _{\mathbf{X}}},f} \right\rangle _\mathcal{H}} \triangleq {\mathbb{E}_{\mathbf{X}}}\left[ {f\left( {\mathbf{X}} \right)} \right],\forall f \in \mathcal{H}$. This kind of kernel mean embedding provides us a nonparametric perspective on manipulating distributions by drawing samples from them. We will require a characteristic kernel $k$ such that the kernel embedding ${\mu_{\mathbf{X}}}\left( P \right) $ is injective, and that the embedding of distributions into infinite-dimensional feature spaces can preserve all of the statistical features of arbitrary distributions, which removes the necessity of density estimation of $P$. This technique has been widely applied in many tasks, including feature extraction, density estimation and two-sample test \cite{cite:ALT07KSE,cite:JMLR12MMD}.

While the true distribution $P(\mathbf{X})$ is rarely accessible, we can estimate
its embedding using a finite sample \cite{cite:JMLR12MMD}. Given a sample ${\mathcal{D}_{\mathbf{X}}} = \{ {{\mathbf{x}}_1}, \ldots ,{{\mathbf{x}}_n}\} $ of size $n$ drawn i.i.d. from $P(\mathbf{X})$, the empirical kernel embedding is
\begin{equation}\label{eqn:Exhat}
	{{\widehat \mu }_{\mathbf{X}}} = \frac{1}{n}\sum\limits_{i = 1}^n {\phi \left( {{{\mathbf{x}}_i}} \right)}.
\end{equation}
This empirical estimate converges to its population counterpart in RKHS norm ${\left\| {{\mu _{\mathbf{X}}} - {{\widehat \mu }_{\mathbf{X}}}} \right\|_\mathcal{H}}$ with a rate of $O(n^{-\frac{1}{2}})$.

Kernel embeddings can be readily generalized to \emph{joint} distributions of two or more variables using tensor product feature spaces \cite{cite:ICML09HSE,cite:ICML10HSE,cite:NIPS13HSE}. A joint distribution $P$ of variables $\mathbf{X}^1, \ldots, \mathbf{X}^m$ can be embedded into an $m$-th order tensor product feature space $ \otimes _{\ell  = 1}^{m}{\mathcal{H}^\ell }$ by
\begin{equation}\label{eqn:Exy}
\begin{aligned}
	{\mathcal{C}_{{{\mathbf{X}}^{1:m}}}} (P) & \triangleq {\mathbb{E}_{{{\mathbf{X}}^{1:m}}}}\left[ { \otimes _{\ell  = 1}^m{\phi ^\ell }\left( {{{\mathbf{X}}^\ell }} \right)} \right] \\
	& = \int\nolimits_{ \times _{\ell  = 1}^m{\Omega ^\ell }} {\left( { \otimes _{\ell  = 1}^m{\phi ^\ell }\left( {\mathbf{x}^\ell} \right)} \right){\text{d}}P\left( {{{\mathbf{x}}^1}, \ldots ,{{\mathbf{x}}^m}} \right)}, 
\end{aligned}
\end{equation}
where ${{\mathbf{X}}^{1:m}}$ denotes the set of $m$ variables $ \{ {{{\mathbf{X}}^1}, \ldots ,{{\mathbf{X}}^m}} \}$ on domain ${ \times _{\ell  = 1}^m{\Omega ^\ell }} = {\Omega ^1} \times  \ldots  \times {\Omega ^m}$, $\phi^\ell$ is the feature map endowed with kernel $k^\ell$ in RKHS $\mathcal{H}^\ell$ for variable $\mathbf{X}^\ell$, $ \otimes _{\ell  = 1}^m{\phi ^\ell }\left( {{{\mathbf{x}}^\ell }} \right) = {\phi ^1}\left( {{{\mathbf{x}}^1}} \right) \otimes  \ldots  \otimes {\phi ^m}\left( {{{\mathbf{x}}^m}} \right)$ is the feature map in the tensor product Hilbert space, where the inner product satisfies $\langle { \otimes _{\ell  = 1}^m{\phi ^\ell }( {{{\mathbf{x}}^\ell }} ), \otimes _{\ell  = 1}^m{\phi ^\ell }( {{{{\mathbf{x'}}}^\ell }} )} \rangle  = \prod\nolimits_{\ell  = 1}^m {{k^\ell }( {{{\mathbf{x}}^\ell },{{{\mathbf{x'}}}^\ell }} )} $. The joint embeddings can be viewed as an uncentered cross-covariance operator $\mathcal{C}_{\mathbf{X}^{1:m}}$ by the standard equivalence between tensor and linear map \cite{cite:ICML10HSE}. That is, given a set of functions $f^1,\ldots,f^m$, their covariance can be computed by ${\mathbb{E}_{{{\mathbf{X}}^{1:m}}}}\left[ {\prod\nolimits_{\ell  = 1}^m {{f^\ell }({{\mathbf{X}}^\ell })} } \right] = \left\langle { \otimes _{\ell  = 1}^m{f^\ell },{\mathcal{C}_{{{\mathbf{X}}^{1:m}}}}} \right\rangle $.

When the true distribution $P(\mathbf{X}^1,\ldots,\mathbf{X}^m)$ is unknown, we can estimate
its embedding using a finite sample \cite{cite:IEEE13KSE}. Given a sample ${\mathcal{D}_{\mathbf{X}^{1:m}}} = \{ {{\mathbf{x}}_1^{1:m}}, \ldots ,{{\mathbf{x}}_n^{1:m}}\} $ of size $n$ drawn i.i.d. from $P(\mathbf{X}^1,\ldots,\mathbf{X}^m)$, the empirical joint embedding (the cross-covariance operator) is estimated as
\begin{equation}\label{eqn:Exyhat}
	{{\widehat{\mathcal{C}}}_{{{\mathbf{X}}^{1:m}}}} = \frac{1}{n}\sum\limits_{i = 1}^n { \otimes _{\ell  = 1}^m{\phi ^\ell }\left( {{\mathbf{x}}_i^\ell } \right)}.
\end{equation}
This empirical estimate converges to its population counterpart with a similar convergence rate as marginal embedding.

\subsection{Maximum Mean Discrepancy}
Let ${\mathcal{D}_{{{\mathbf{X}}^s}}} = \{ {\mathbf{x}}_1^s, \ldots ,{\mathbf{x}}_{{n_s}}^s\} $ and ${\mathcal{D}_{{{\mathbf{X}}^t}}} = \{ {\mathbf{x}}_1^t, \ldots ,{\mathbf{x}}_{{n_t}}^t\} $ be the sets of samples from distributions $P(\mathbf{X}^s)$ and $Q(\mathbf{X}^t)$, respectively. Maximum Mean Discrepancy (MMD) \cite{cite:JMLR12MMD} is a kernel two-sample test which rejects or accepts the null hypothesis $P = Q$ based on the observed samples. The basic idea behind MMD is that if the generating distributions are identical, all the statistics are the same. Formally, MMD defines the following difference measure:
\begin{equation}\label{eqn:MMD}
	{D_\mathcal{H}}\left( {P,Q} \right) \triangleq \mathop {\sup }\limits_{f \in \mathcal{H}} \left( {{\mathbb{E}_{{{\mathbf{X}}^s}}}\left[ {f\left( {{{\mathbf{X}}^s}} \right)} \right] - {\mathbb{E}_{{{\mathbf{X}}^t}}}\left[ {f\left( {{{\mathbf{X}}^t}} \right)} \right]} \right),
\end{equation}
where $\mathcal{H}$ is a class of functions. It is shown that the class of functions in an universal RKHS $\mathcal{H}$ is rich enough to distinguish any two distributions and MMD is expressed as the distance between their mean embeddings: ${D_\mathcal{H}}\left( {P,Q} \right) = \left\| {{\mu _{{{\mathbf{X}}^s}}}\left( P \right) - {\mu _{{{\mathbf{X}}^t}}}\left( Q \right)} \right\|_\mathcal{H}^2$. The main theoretical result is that $P=Q$ if and only if ${D_\mathcal{H}}\left( {P,Q} \right) = 0$ \cite{cite:JMLR12MMD}. 

In practice, an estimate of the MMD compares the square distance between the empirical kernel mean embeddings as
\begin{equation}\label{eqn:MMDhat}
\begin{aligned}
  {{\widehat D}_\mathcal{H}}\left( {P,Q} \right) &= \frac{1}{{n_s^2}}\sum\limits_{i = 1}^{{n_s}} {\sum\limits_{j = 1}^{{n_s}} {k\left( {{\mathbf{x}}_i^s,{\mathbf{x}}_j^s} \right)} }  \\
   & + \frac{1}{{n_t^2}}\sum\limits_{i = 1}^{{n_t}} {\sum\limits_{j = 1}^{{n_t}} {k\left( {{\mathbf{x}}_i^t,{\mathbf{x}}_j^t} \right)} } \\
   & - \frac{2}{{{n_s}{n_t}}}\sum\limits_{i = 1}^{{n_s}} {\sum\limits_{j = 1}^{{n_t}} {k\left( {{\mathbf{x}}_i^s,{\mathbf{x}}_j^t} \right)} }, \\
\end{aligned}
\end{equation}
where ${{\widehat D}_\mathcal{H}}\left( {P,Q} \right)$ is an unbiased estimator of ${{D}_\mathcal{H}}\left( {P,Q} \right)$.

\begin{figure*}[!htb]
  \centering
  \subfigure[Joint Adaptation Network (JAN)]{
    \includegraphics[width=0.48\textwidth]{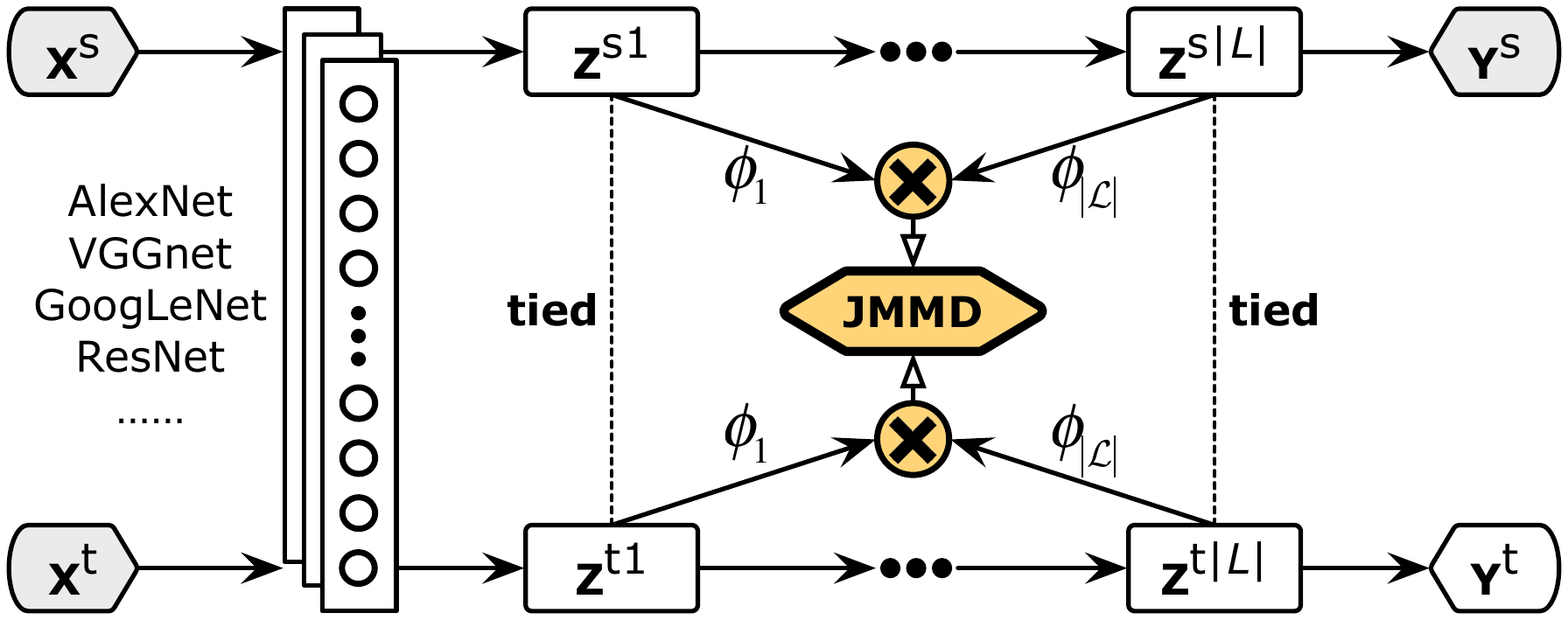}
    \label{fig:JANa}
  }
  \subfigure[Adversarial Joint Adaptation Network (JAN-A)]{
    \includegraphics[width=0.48\textwidth]{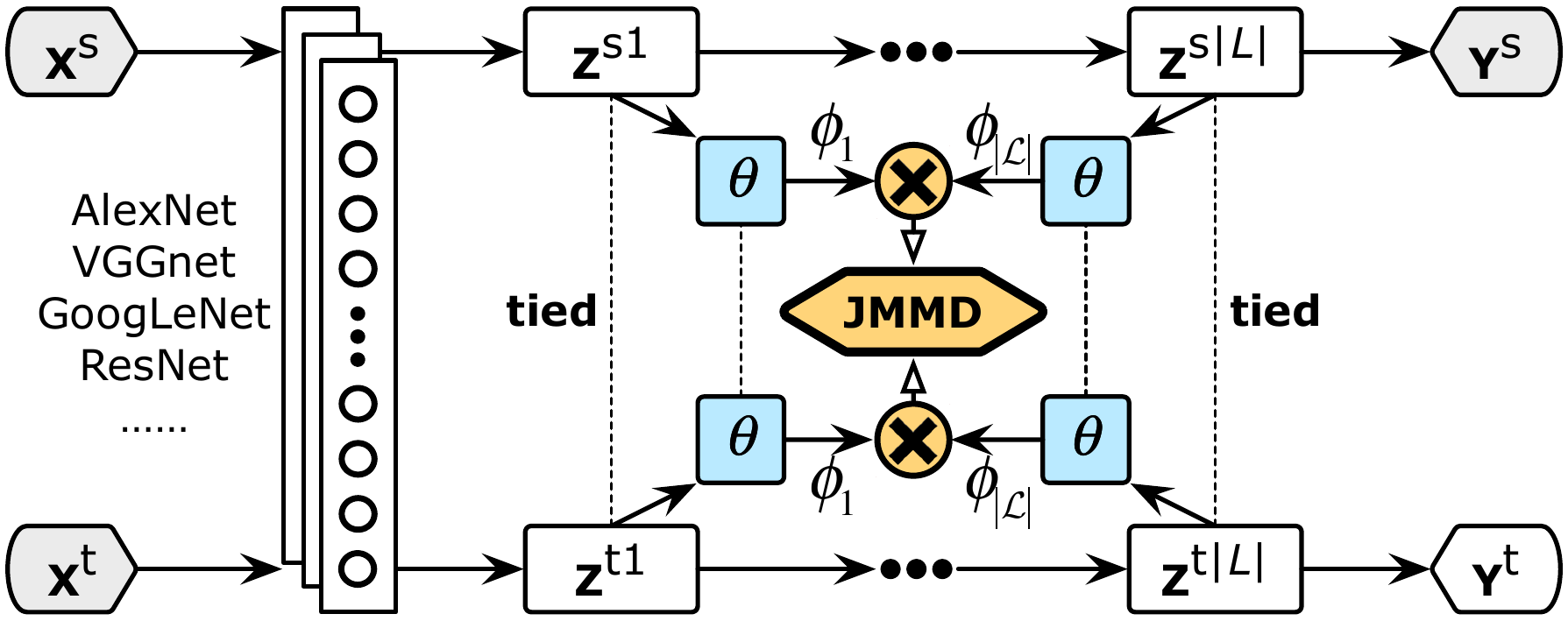}
    \label{fig:JANb}
  }
  \vspace{-10pt}
  \caption{The architectures of Joint Adaptation Network (JAN) (a) and its adversarial version (JAN-A) (b). Since deep features eventually transition from general to specific along the network, activations in multiple domain-specific layers $\mathcal{L}$ are not safely transferable. And the joint distributions of the activations  $P(\mathbf{Z}^{s1},\ldots,\mathbf{Z}^{s|\mathcal{L}|})$ and $Q(\mathbf{Z}^{t1},\ldots,\mathbf{Z}^{t|\mathcal{L}|})$ in these layers should be adapted by JMMD minimization.}
  \label{fig:JAN}
  \vspace{-10pt}
\end{figure*}

\section{Joint Adaptation Networks}
In unsupervised domain adaptation, we are given a source domain $\mathcal{D}_s = \{(\mathbf{x}_i^s,{\bf y}^s_i)\}_{i=1}^{n_s}$ of $n_s$ labeled examples and a target domain ${{\cal D}_t} = \{ {\bf{x}}_j^t\} _{j = 1}^{{n_t}}$ of $n_t$ unlabeled examples. The source domain and target domain are sampled from joint distributions $P(\mathbf{X}^s, \mathbf{Y}^s)$ and $Q(\mathbf{X}^t, \mathbf{Y}^t)$ respectively, $P \ne Q$. The goal of this paper is to design a deep neural network ${\bf y} = f\left( {\bf{x}} \right)$ which formally reduces the shifts in the joint distributions across domains and enables learning both transferable features and classifiers, such that the target risk ${R_t}\left( f \right) = {\mathbb{E} _{\left( {{\mathbf{x}},{\bf y}} \right) \sim Q}}\left[ {f \left( {\mathbf{x}} \right) \ne {\bf y}} \right]$ can be minimized by jointly minimizing the source risk and domain discrepancy.

Recent studies reveal that deep networks \cite{cite:TPAMI13DLSurvey} can learn more transferable representations than traditional hand-crafted features \cite{cite:CVPR13MidLevel,cite:NIPS14CNN}. The favorable transferability of deep features leads to several state of the art deep transfer learning methods \cite{cite:ICML15RevGrad,cite:ICCV15SDT,cite:ICML15DAN,cite:NIPS16RTN}. This paper also tackles unsupervised domain adaptation by learning transferable features using deep neural networks. We extend deep convolutional neural networks (CNNs), including AlexNet \cite{cite:NIPS12CNN} and ResNet \cite{cite:CVPR16DRL}, to novel joint adaptation networks (JANs) as shown in Figure~\ref{fig:JAN}. The empirical error of CNN classifier $f({\bf x})$ on source domain labeled data $\mathcal{D}_s$ is
\begin{equation}\label{eqn:CNN}
	\mathop {\min }\limits_{f} \frac{1}{{{n_s}}}\sum\limits_{i = 1}^{{n_s}} {J\left( {f\left( {{\bf{x}}_i^s} \right),{\bf y}_i^s} \right)} ,
\end{equation}
where $J(\cdot,\cdot)$ is the cross-entropy loss function. Based on the quantification study of feature transferability in deep convolutional networks \cite{cite:NIPS14CNN}, convolutional layers can learn generic features that are transferable across domains \cite{cite:NIPS14CNN}. Thus we opt to fine-tune the features of convolutional layers when transferring pre-trained deep models from source domain to target domain.

However, the literature findings also reveal that the deep features can reduce, but not remove, the cross-domain distribution discrepancy \cite{cite:NIPS14CNN,cite:ICML15DAN,cite:NIPS16RTN}. The deep features in standard CNNs must eventually transition from general to specific along the network, and the transferability of features and classifiers decreases when the cross-domain discrepancy increases \cite{cite:NIPS14CNN}. In other words, even feed-forwarding the source and target domain data through the deep network for multilayer feature abstraction, the shifts in the joint distributions $P(\mathbf{X}^s, \mathbf{Y}^s)$ and $Q(\mathbf{X}^t, \mathbf{Y}^t)$ still linger in the activations ${\mathbf{Z}^1},\ldots,{\mathbf{Z}^{|\mathcal{L}|}}$ of the higher network layers $\mathcal{L}$. Taking AlexNet \cite{cite:NIPS12CNN} as an example, the activations in the higher fully-connected layers $\mathcal{L} = \{ fc6, fc7, fc8\}$ are not safely transferable for domain adaptation \cite{cite:NIPS14CNN}. Note that the shift in the feature distributions $P(\mathbf{X}^s)$ and $Q(\mathbf{X}^t)$ mainly lingers in the feature layers $fc6, fc7$ while the shift in the label distributions $P(\mathbf{Y}^s)$ and $Q(\mathbf{Y}^t)$ mainly lingers in the classifier layer $fc8$. Thus we can use the joint distributions of the activations in layers $\mathcal{L}$, i.e. $P(\mathbf{Z}^{s1},\ldots,\mathbf{Z}^{s|\mathcal{L}|})$ and $Q(\mathbf{Z}^{t1},\ldots,\mathbf{Z}^{t|\mathcal{L}|})$ as good surrogates of the original joint distributions $P(\mathbf{X}^s, \mathbf{Y}^s)$ and $Q(\mathbf{X}^t, \mathbf{Y}^t)$, respectively. To enable unsupervised domain adaptation, we should find a way to match $P(\mathbf{Z}^{s1},\ldots,\mathbf{Z}^{s|\mathcal{L}|})$ and $Q(\mathbf{Z}^{t1},\ldots,\mathbf{Z}^{t|\mathcal{L}|})$.

\subsection{Joint Maximum Mean Discrepancy}
Many existing methods address transfer learning by bounding the target error with the source error plus a discrepancy between the marginal distributions $P({\mathbf{X}^s})$ and $Q({\mathbf{X}^t})$ of the source and target domains \cite{cite:ML10DAT}. The Maximum Mean Discrepancy (MMD) \cite{cite:JMLR12MMD}, as a kernel two-sample test statistic, has been widely applied to measure the discrepancy in marginal distributions $P(\mathbf{X}^s)$ and $Q(\mathbf{X}^t)$ \cite{cite:Arxiv14DDC,cite:ICML15DAN,cite:NIPS16RTN}. To date MMD has not been used to measure the discrepancy in joint distributions $P(\mathbf{Z}^{s1},\ldots,\mathbf{Z}^{s|\mathcal{L}|})$ and $Q(\mathbf{Z}^{t1},\ldots,\mathbf{Z}^{t|\mathcal{L}|})$, possibly because MMD has not been directly defined for joint distributions by \cite{cite:JMLR12MMD} while in conventional shallow domain adaptation methods the joint distributions are not easy to manipulate and match.

Following the virtue of MMD \eqref{eqn:MMD}, we use the Hilbert space embeddings of joint distributions \eqref{eqn:Exy} to measure the discrepancy of two joint distributions $P(\mathbf{Z}^{s1},\ldots,\mathbf{Z}^{s|\mathcal{L}|})$ and $Q(\mathbf{Z}^{t1},\ldots,\mathbf{Z}^{t|\mathcal{L}|})$. The resulting measure is called Joint Maximum Mean Discrepancy (JMMD), which is defined as
\begin{equation}\label{eqn:JMMD}
	{D_\mathcal{L}}\left( {P,Q} \right) \triangleq \left\| {{\mathcal{C}_{{{\mathbf{Z}}^{s,1:|\mathcal{L}|}}}}\left( P \right) - {\mathcal{C}_{{{\mathbf{Z}}^{t,1:|\mathcal{L}|}}}}\left( Q \right)} \right\|_{ \otimes _{\ell  = 1}^{|\mathcal{L}|}{\mathcal{H}^\ell }}^2.
\end{equation}
Based on the virtue of the kernel two-sample test theory \cite{cite:JMLR12MMD}, we will have $P(\mathbf{Z}^{s1},\ldots,\mathbf{Z}^{s|\mathcal{L}|}) = Q(\mathbf{Z}^{t1},\ldots,\mathbf{Z}^{t|\mathcal{L}|})$ if and only if $D_{\mathcal{L}}(P,Q)=0$. Given source domain $\mathcal{D}_s$ of $n_s$ labeled points and target domain ${{\cal D}_t}$ of $n_t$ unlabeled points drawn i.i.d. from $P$ and $Q$ respectively, the deep networks will generate activations in layers $\mathcal{L}$ as $\{(\mathbf{z}_i^{s1},\ldots,{\bf z}^{s|\mathcal{L}|}_{i})\}_{i=1}^{n_s}$ and $\{ ({\bf{z}}_j^{t1},\ldots,{\bf z}_j^{t|\mathcal{L}|})\} _{j = 1}^{{n_t}}$. The empirical estimate of $D_{\mathcal{L}}(P,Q)$ is computed as the squared distance between the empirical kernel mean embeddings as
\begin{equation}\label{eqn:JMMDhat}
  \begin{aligned}
  {{\widehat{D}}_\mathcal{L}}\left( {P,Q} \right) &= \frac{1}{{n_s^2}}\sum\limits_{i = 1}^{{n_s}} {\sum\limits_{j = 1}^{{n_s}} {\prod\limits_{\ell  \in \mathcal{L}} {{k^\ell }\left( {{\mathbf{z}}_i^{s\ell },{\mathbf{z}}_j^{s\ell }} \right)} } }   \\
   &+ \frac{1}{{n_t^2}}\sum\limits_{i = 1}^{{n_t}} {\sum\limits_{j = 1}^{{n_t}} {\prod\limits_{\ell  \in \mathcal{L}} {{k^\ell }\left( {{\mathbf{z}}_i^{t\ell },{\mathbf{z}}_j^{t\ell }} \right)} } }   \\
   &- \frac{2}{{{n_s}{n_t}}}\sum\limits_{i = 1}^{{n_s}} {\sum\limits_{j = 1}^{{n_t}} {\prod\limits_{\ell  \in \mathcal{L}} {{k^\ell }\left( {{\mathbf{z}}_i^{s\ell },{\mathbf{z}}_j^{t\ell }} \right)} } }.   \\ 
  \end{aligned}
\end{equation}
\textbf{Remark:} Taking a close look on the objectives of MMD \eqref{eqn:MMDhat} and JMMD \eqref{eqn:JMMDhat}, we can find some interesting connections. The difference is that, for the activations ${\mathbf{Z}^\ell}$ in each layer $\ell\in\mathcal{L}$, instead of putting uniform weights on the kernel function $k^\ell(\mathbf{z}_i^\ell,\mathbf{z}_j^\ell)$ as in MMD, JMMD applies non-uniform weights, reflecting the influence of other variables in other layers $\mathcal{L}\backslash \ell $. This captures the full interactions between different variables in the joint distributions $P(\mathbf{Z}^{s1},\ldots,\mathbf{Z}^{s|\mathcal{L}|})$ and $Q(\mathbf{Z}^{t1},\ldots,\mathbf{Z}^{t|\mathcal{L}|})$, which is crucial for domain adaptation. All previous deep transfer learning methods \cite{cite:Arxiv14DDC,cite:ICML15DAN,cite:ICML15RevGrad,cite:ICCV15SDT,cite:NIPS16RTN} have not addressed this issue.

\subsection{Joint Adaptation Networks}
Denote by $\mathcal{L}$ the domain-specific layers where the activations are not safely transferable. We will formally reduce the discrepancy in the joint distributions of the activations in layers $\mathcal{L}$, i.e. $P(\mathbf{Z}^{s1},\ldots,\mathbf{Z}^{s|\mathcal{L}|})$ and $Q(\mathbf{Z}^{t1},\ldots,\mathbf{Z}^{t|\mathcal{L}|})$. Note that the features in the lower layers of the network are transferable and hence will not require a further distribution matching. By integrating the JMMD \eqref{eqn:JMMDhat} over the domain-specific layers $\mathcal{L}$ into the CNN error \eqref{eqn:CNN}, the joint distributions are matched end-to-end with network training, 
\begin{equation}
  \mathop {\min }\limits_f \frac{1}{{{n_s}}}\sum\limits_{i = 1}^{{n_s}} {J\left( {f\left( {{\mathbf{x}}_i^s} \right),{\mathbf{y}}_i^s} \right)}  + \lambda {{\widehat D}_\mathcal{L}}\left( P, Q \right),
\end{equation}
where $\lambda>0$ is a tradeoff parameter of the JMMD penalty. As shown in Figure~\ref{fig:JANa}, we set ${\cal L} = \{fc6, fc7, fc8\}$ for the JAN model based on AlexNet (last three layers) while we set ${\cal L} = \{pool5, fc\}$ for the JAN model based on ResNet (last two layers), as these layers are tailored to task-specific structures, which are not safely transferable and should be jointly adapted by minimizing CNN error and JMMD \eqref{eqn:JMMDhat}.

A limitation of JMMD \eqref{eqn:JMMDhat} is its quadratic complexity, which is inefficient for scalable deep transfer learning. Motivated by the unbiased estimate of MMD \cite{cite:JMLR12MMD}, we derive a similar linear-time estimate of JMMD as follows,
\begin{equation}
\small
\begin{aligned}
  {{\widehat D}_\mathcal{L}}\left( {P,Q} \right) & = {\frac{2}{n}\sum\limits_{i = 1}^{n/2} {\left( {\prod\limits_{\ell  \in \mathcal{L}} {{k^\ell }( {{\mathbf{z}}_{2i - 1}^{s\ell },{\mathbf{z}}_{2i}^{s\ell }} )}  + \prod\limits_{\ell  \in \mathcal{L}} {{k^\ell }( {{\mathbf{z}}_{2i - 1}^{t\ell },{\mathbf{z}}_{2i}^{t\ell }} )} } \right)}}  \\
   & - {\frac{2}{n}\sum\limits_{i = 1}^{n/2} {\left( {\prod\limits_{\ell  \in \mathcal{L}} {{k^\ell }( {{\mathbf{z}}_{2i - 1}^{s\ell },{\mathbf{z}}_{2i}^{t\ell }} )}  + \prod\limits_{\ell  \in \mathcal{L}} {{k^\ell }( {{\mathbf{z}}_{2i - 1}^{t\ell },{\mathbf{z}}_{2i}^{s\ell }} )} } \right)}}, \\ 
\end{aligned}
\end{equation}
where $n=n_s$. This linear-time estimate well fits the mini-batch stochastic gradient descent (SGD) algorithm. In each mini-batch, we sample the same number of source points and target points to eliminate the bias caused by domain size. This enables our models to scale linearly to large samples.

\subsection{Adversarial Training for Optimal MMD}
The MMD defined using the RKHS \eqref{eqn:MMDhat} has the advantage of not requiring a separate network to approximately maximize the original definition of MMD \eqref{eqn:MMD}. But the original MMD \eqref{eqn:MMD} reveals that, in order to maximize the test power such that any two distributions can be distinguishable, we require the class of functions $f \in \mathcal{H}$ to be rich enough. Although \cite{cite:JMLR12MMD} shows that an universal RKHS is rich enough, such kernel-based MMD may suffer from vanishing gradients for low-bandwidth kernels. Moreover, it may be possible that some widely-used kernels are unable to capture very complex distances in high dimensional spaces such as natural images \cite{cite:AIS15MMD,cite:Arxiv17WGAN}. 

To circumvent the issues of vanishing gradients and non-rich function class of kernel-based MMD \eqref{eqn:MMDhat}, we are enlightened by the original MMD \eqref{eqn:MMD} which fits the adversarial training in GANs \cite{cite:NIPS14AdversarialNet}. We add multiple fully-connected layers parametrized by $\theta$ to the proposed JMMD \eqref{eqn:JMMDhat} to make the function class of JMMD richer using neural network as shown in Figure~\ref{fig:JANb}. We maximize JMMD with respect to these new parameters $\theta$ to approach the virtue of the original MMD \eqref{eqn:MMD}, that is, maximizing the test power of JMMD such that distributions of source and target domains are made more distinguishable \cite{cite:NIPS09KCC}. This leads to a new adversarial joint adaptation network as
\begin{equation}
	\mathop {\min }\limits_f \mathop {\max }\limits_\theta  \frac{1}{{{n_s}}}\sum\limits_{i = 1}^{{n_s}} {J\left( {f\left( {{\mathbf{x}}_i^s} \right),{\mathbf{y}}_i^s} \right)}  + \lambda {{\widehat D}_\mathcal{L}}\left( {P,Q;\theta } \right).
\end{equation}
Learning deep features by minimizing this more powerful JMMD, intuitively any shift in the joint distributions will be more easily identified by JMMD and then adapted by CNN.

\textbf{Remark:} This version of JAN shares the idea of domain-adversarial training with \cite{cite:ICML15RevGrad}, but differs in that we use the JMMD as the domain adversary while \cite{cite:ICML15RevGrad} uses logistic regression. As pointed out in a very recent study \cite{cite:Arxiv17WGAN}, our JMMD-adversarial network can be trained more easily.

\section{Experiments}
We evaluate the joint adaptation networks with state of the art transfer learning and deep learning methods. Codes and datasets are available at \url{http://github.com/thuml}.

\subsection{Setup}
\textbf{Office-31} \cite{cite:ECCV10Office} is a standard benchmark for domain adaptation in computer vision, comprising 4,652 images and 31 categories collected from three distinct domains: \textit{Amazon} (\textbf{A}), which contains images downloaded from \url{amazon.com}, \textit{Webcam} (\textbf{W}) and \textit{DSLR} (\textbf{D}), which contain images respectively taken by web camera and digital SLR camera under different settings. We evaluate all methods across three transfer tasks \textbf{A} $\rightarrow$ \textbf{W}, \textbf{D} $\rightarrow$ \textbf{W} and \textbf{W} $\rightarrow$ \textbf{D}, which are widely adopted by previous deep transfer learning methods \cite{cite:Arxiv14DDC,cite:ICML15RevGrad}, and another three transfer tasks \textbf{A} $\rightarrow$ \textbf{D}, \textbf{D} $\rightarrow$ \textbf{A} and \textbf{W} $\rightarrow$ \textbf{A} as in \cite{cite:ICML15DAN,cite:NIPS16RTN,cite:ICCV15SDT}.

\textbf{ImageCLEF-DA}\footnote{\url{http://imageclef.org/2014/adaptation}} is a benchmark dataset for ImageCLEF 2014 domain adaptation challenge, which is organized by selecting the 12 common categories shared by the following three public datasets, each is considered as a domain: \textit{Caltech-256} (\textbf{C}), \textit{ImageNet ILSVRC 2012} (\textbf{I}), and \textit{Pascal VOC 2012} (\textbf{P}). 
There are 50 images in each category and 600 images in each domain. We use all domain combinations and build 6 transfer tasks: \textbf{I} $\rightarrow$ \textbf{P}, \textbf{P} $\rightarrow$ \textbf{I}, \textbf{I} $\rightarrow$ \textbf{C}, \textbf{C} $\rightarrow$ \textbf{I}, \textbf{C} $\rightarrow$ \textbf{P}, and \textbf{P} $\rightarrow$ \textbf{C}. Different from \emph{Office-31} where different domains are of different sizes, the three domains in ImageCLEF-DA are of equal size, which makes it a good complement to  \textit{Office-31} for more controlled experiments.

We compare with conventional and state of the art transfer learning and deep learning methods: Transfer Component Analysis (\textbf{TCA}) \cite{cite:TNN11TCA}, Geodesic Flow Kernel (\textbf{GFK}) \cite{cite:CVPR12GFK}, Convolutional Neural Networks \textbf{AlexNet} \cite{cite:NIPS12CNN} and \textbf{ResNet} \cite{cite:CVPR16DRL}, Deep Domain Confusion (\textbf{DDC}) \cite{cite:Arxiv14DDC}, Deep Adaptation Network (\textbf{DAN}) \cite{cite:ICML15DAN}, Reverse Gradient (\textbf{RevGrad}) \cite{cite:ICML15RevGrad}, and Residual Transfer Network (\textbf{RTN}) \cite{cite:NIPS16RTN}. TCA is a transfer learning method based on MMD-regularized Kernel PCA. GFK is a manifold learning method that interpolates across an infinite number of intermediate subspaces to bridge domains. DDC is the first method that maximizes domain invariance by regularizing the adaptation layer of AlexNet using linear-kernel MMD \cite{cite:JMLR12MMD}. DAN learns transferable features by embedding deep features of multiple task-specific layers to reproducing kernel Hilbert spaces (RKHSs) and matching different distributions optimally using multi-kernel MMD. RevGrad improves domain adaptation by making the source and target domains indistinguishable for a domain discriminator by adversarial training. RTN jointly learns transferable features and adaptive classifiers by deep residual learning \cite{cite:CVPR16DRL}.

We examine the influence of deep representations for domain adaptation by employing the breakthrough \textbf{AlexNet} \cite{cite:NIPS12CNN} and the state of the art \textbf{ResNet} \cite{cite:CVPR16DRL} for learning transferable deep representations. For AlexNet, we follow DeCAF \cite{cite:ICML14DeCAF} and use the activations of layer $fc7$  as image representation. For ResNet (50 layers), we use the activations of the last feature layer $pool5$ as image representation. We follow standard evaluation protocols for unsupervised domain adaptation \cite{cite:ICML15DAN,cite:ICML15RevGrad}. For both \emph{Office-31} and \emph{ImageCLEF-DA} datasets, we use all labeled source examples and all unlabeled target examples. We compare the average classification accuracy of each method on three random experiments, and report the standard error of the classification accuracies by different experiments of the same transfer task. We perform model selection by tuning hyper-parameters using transfer cross-validation \cite{cite:ECML10TCV}. For MMD-based methods and JAN, we adopt Gaussian kernel with bandwidth set to median pairwise squared distances on the training data \cite{cite:JMLR12MMD}.

We implement all deep methods based on the \textbf{Caffe} framework, and fine-tune from Caffe-provided models of AlexNet \cite{cite:NIPS12CNN} and ResNet \cite{cite:CVPR16DRL}, both are pre-trained on the ImageNet 2012 dataset. We fine-tune all convolutional and pooling layers and train the classifier layer via back propagation. Since the classifier is trained from scratch, we set its learning rate to be 10 times that of the other layers. We use mini-batch stochastic gradient descent (SGD) with momentum of 0.9 and the learning rate annealing strategy in RevGrad \cite{cite:ICML15RevGrad}: the learning rate is not selected by a grid search due to high computational cost---it is adjusted during SGD using the following formula: ${\eta _p} = \frac{{{\eta _0}}}{{{{\left( {1 + \alpha p} \right)}^\beta }}}$, where $p$ is the training progress linearly changing from $0$ to $1$, $\eta_0 = 0.01, \alpha=10$ and $\beta=0.75$, which is optimized to promote convergence and low error on the source domain. To suppress noisy activations at the early stages of training, instead of fixing the adaptation factor $\lambda$, we gradually change it from $0$ to $1$ by a progressive schedule: ${\lambda _p} = \frac{2}{{1 + \exp \left( { - \gamma p} \right)}} - 1$, and $\gamma = 10$ is fixed throughout experiments \cite{cite:ICML15RevGrad}. This progressive strategy significantly stabilizes parameter sensitivity and eases model selection for JAN and JAN-A.

\begin{table*}[!htbp]
  \addtolength{\tabcolsep}{2.0pt}
  \centering
  \caption{Classification accuracy (\%) on \emph{Office-31} dataset for unsupervised domain adaptation (AlexNet and ResNet)}
  \label{table:office31}
  \begin{small}
  \begin{tabular}{cccccccc}
    \Xhline{1pt}
    Method & A $\rightarrow$ W & D $\rightarrow$ W & W $\rightarrow$ D & A $\rightarrow$ D & D $\rightarrow$ A & W $\rightarrow$ A & Avg \\
    \hline
    AlexNet \cite{cite:NIPS12CNN} & 61.6$\pm$0.5 & 95.4$\pm$0.3 & 99.0$\pm$0.2 & 63.8$\pm$0.5 & 51.1$\pm$0.6 & 49.8$\pm$0.4 & 70.1 \\
    TCA \cite{cite:TNN11TCA} & 61.0$\pm$0.0 & 93.2$\pm$0.0 & 95.2$\pm$0.0 & 60.8$\pm$0.0 & 51.6$\pm$0.0 & 50.9$\pm$0.0 & 68.8 \\
    GFK \cite{cite:CVPR12GFK} & 60.4$\pm$0.0 & 95.6$\pm$0.0 & 95.0$\pm$0.0 & 60.6$\pm$0.0 & 52.4$\pm$0.0 & 48.1$\pm$0.0 & 68.7 \\
    DDC \cite{cite:Arxiv14DDC} & 61.8$\pm$0.4 & 95.0$\pm$0.5 & 98.5$\pm$0.4 & 64.4$\pm$0.3 & 52.1$\pm$0.6 & 52.2$\pm$0.4 & 70.6 \\
    DAN \cite{cite:ICML15DAN} & 68.5$\pm$0.5 & 96.0$\pm$0.3 & 99.0$\pm$0.3 & 67.0$\pm$0.4 & 54.0$\pm$0.5 & 53.1$\pm$0.5 & 72.9 \\
    RTN \cite{cite:NIPS16RTN} & 73.3$\pm$0.3 & \textbf{96.8}$\pm$0.2 & \textbf{99.6}$\pm$0.1 & 71.0$\pm$0.2 & 50.5$\pm$0.3 & 51.0$\pm$0.1 & 73.7 \\
    RevGrad \cite{cite:ICML15RevGrad} & 73.0$\pm$0.5 & 96.4$\pm$0.3 & 99.2$\pm$0.3 & 72.3$\pm$0.3 & 53.4$\pm$0.4 & 51.2$\pm$0.5 & 74.3 \\
    JAN (ours) & 74.9$\pm$0.3 & 96.6$\pm$0.2 & 99.5$\pm$0.2 & 71.8$\pm$0.2 & \textbf{58.3}$\pm$0.3 & 55.0$\pm$0.4 & 76.0 \\
    JAN-A (ours) & \textbf{75.2}$\pm$0.4 & 96.6$\pm$0.2 & \textbf{99.6}$\pm$0.1 & \textbf{72.8}$\pm$0.3 & 57.5$\pm$0.2 & \textbf{56.3}$\pm$0.2 & \textbf{76.3} \\
    \hline
    ResNet \cite{cite:CVPR16DRL} & 68.4$\pm$0.2 & 96.7$\pm$0.1 & 99.3$\pm$0.1 & 68.9$\pm$0.2 & 62.5$\pm$0.3 & 60.7$\pm$0.3 & 76.1 \\
    TCA \cite{cite:TNN11TCA} & 72.7$\pm$0.0 & 96.7$\pm$0.0 & 99.6$\pm$0.0 & 74.1$\pm$0.0 & 61.7$\pm$0.0 & 60.9$\pm$0.0 & 77.6 \\
    GFK \cite{cite:CVPR12GFK} & 72.8$\pm$0.0 & 95.0$\pm$0.0 & 98.2$\pm$0.0 & 74.5$\pm$0.0 & 63.4$\pm$0.0 & 61.0$\pm$0.0 & 77.5 \\
    DDC \cite{cite:Arxiv14DDC} & 75.6$\pm$0.2 & 96.0$\pm$0.2 & 98.2$\pm$0.1 & 76.5$\pm$0.3 & 62.2$\pm$0.4 & 61.5$\pm$0.5 & 78.3 \\
    DAN \cite{cite:ICML15DAN} & 80.5$\pm$0.4 & 97.1$\pm$0.2 & 99.6$\pm$0.1 & 78.6$\pm$0.2 & 63.6$\pm$0.3 & 62.8$\pm$0.2 & 80.4 \\
    RTN \cite{cite:NIPS16RTN} & 84.5$\pm$0.2 & 96.8$\pm$0.1 & 99.4$\pm$0.1 & 77.5$\pm$0.3 & 66.2$\pm$0.2 & 64.8$\pm$0.3 & 81.6 \\
    RevGrad \cite{cite:ICML15RevGrad} & 82.0$\pm$0.4 & 96.9$\pm$0.2 & 99.1$\pm$0.1 & 79.7$\pm$0.4 & 68.2$\pm$0.4 & 67.4$\pm$0.5 & 82.2 \\
    JAN (ours) & 85.4$\pm$0.3 & \textbf{97.4}$\pm$0.2 & \textbf{99.8}$\pm$0.2 & 84.7$\pm$0.3 & 68.6$\pm$0.3 & 70.0$\pm$0.4 & 84.3 \\
    JAN-A (ours) & \textbf{86.0}$\pm$0.4 & 96.7$\pm$0.3 & 99.7$\pm$0.1 & \textbf{85.1}$\pm$0.4 & \textbf{69.2}$\pm$0.4 & \textbf{70.7}$\pm$0.5 & \textbf{84.6} \\
    \Xhline{1pt}
  \end{tabular}
  \end{small}
  \vspace{-5pt}
\end{table*}

\begin{table*}[!htbp]
  \addtolength{\tabcolsep}{2.8pt}
  \centering
  \caption{Classification accuracy (\%) on \emph{ImageCLEF-DA} for unsupervised domain adaptation (AlexNet and ResNet)}
  \label{table:imageclef-da}
  \begin{small}
  \begin{tabular}{cccccccc}
    \Xhline{1pt}
    Method & I $\rightarrow$ P & P $\rightarrow$ I & I $\rightarrow$ C & C $\rightarrow$ I & C $\rightarrow$ P & P $\rightarrow$ C & Avg \\
    \hline
    AlexNet \cite{cite:NIPS12CNN} & 66.2$\pm$0.2 & 70.0$\pm$0.2 & 84.3$\pm$0.2 & 71.3$\pm$0.4 & 59.3$\pm$0.5 & 84.5$\pm$0.3 & 73.9 \\
    DAN \cite{cite:ICML15DAN} & 67.3$\pm$0.2 & 80.5$\pm$0.3 & 87.7$\pm$0.3 & 76.0$\pm$0.3 & 61.6$\pm$0.3 & 88.4$\pm$0.2 & 76.9 \\
    RTN \cite{cite:NIPS16RTN} & \textbf{67.4}$\pm$0.3 & 81.3$\pm$0.3 & 89.5$\pm$0.4 & 78.0$\pm$0.2 & 62.0$\pm$0.2 & 89.1$\pm$0.1 & 77.9 \\
    JAN (ours) & 67.2$\pm$0.5 & \textbf{82.8}$\pm$0.4 & \textbf{91.3}$\pm$0.5 & \textbf{80.0}$\pm$0.5 & \textbf{63.5}$\pm$0.4 & \textbf{91.0}$\pm$0.4 & \textbf{79.3} \\
    \hline
    ResNet \cite{cite:CVPR16DRL} & 74.8$\pm$0.3 & 83.9$\pm$0.1 & 91.5$\pm$0.3 & 78.0$\pm$0.2 & 65.5$\pm$0.3 & 91.2$\pm$0.3 & 80.7 \\
    DAN \cite{cite:ICML15DAN} & 74.5$\pm$0.4 & 82.2$\pm$0.2 & 92.8$\pm$0.2 & 86.3$\pm$0.4 & 69.2$\pm$0.4 & 89.8$\pm$0.4 & 82.5 \\
    RTN \cite{cite:NIPS16RTN} & 74.6$\pm$0.3 & 85.8$\pm$0.1 & 94.3$\pm$0.1 & 85.9$\pm$0.3 & 71.7$\pm$0.3 & 91.2$\pm$0.4 & 83.9 \\
    JAN (ours) & \textbf{76.8}$\pm$0.4 & \textbf{88.0}$\pm$0.2 & \textbf{94.7}$\pm$0.2 & \textbf{89.5}$\pm$0.3 & \textbf{74.2}$\pm$0.3 & \textbf{91.7}$\pm$0.3 & \textbf{85.8} \\
    \Xhline{1pt}
  \end{tabular}
  \end{small}
  \vspace{-5pt}
\end{table*}

\subsection{Results}
The classification accuracy results on the \textit{Office-31} dataset for unsupervised domain adaptation based on AlexNet and ResNet are shown in Table \ref{table:office31}. As fair comparison with identical evaluation setting, the results of DAN \cite{cite:ICML15DAN}, RevGrad \cite{cite:ICML15RevGrad}, and RTN \cite{cite:NIPS16RTN} are directly reported from their published papers. The proposed JAN models outperform all comparison methods on most transfer tasks. It is noteworthy that JANs promote the classification accuracies substantially on hard transfer tasks, e.g. \textbf{D $\rightarrow$ A} and \textbf{W $\rightarrow$ A}, where the source and target domains are substantially different and the source domain is smaller than the target domain, and produce comparable classification accuracies on easy transfer tasks, \textbf{D $\rightarrow$ W} and \textbf{W $\rightarrow$ D}, where the source and target domains are similar \cite{cite:ECCV10Office}. The encouraging results highlight the key importance of joint distribution adaptation in deep neural networks, and suggest that JANs are able to learn more transferable representations for effective domain adaptation.

The results reveal several interesting observations. \textbf{(1)} Standard deep learning methods either outperform (AlexNet) or underperform (ResNet) traditional shallow transfer learning methods (TCA and GFK) using deep features (AlexNet-fc7 and ResNet-pool5) as input. And traditional shallow transfer learning methods perform better with more transferable deep features extracted by ResNet. This confirms the current practice that deep networks learn abstract feature representations, which can only reduce, but not remove, the domain discrepancy \cite{cite:NIPS14CNN}. \textbf{(2)} Deep transfer learning methods substantially outperform both standard deep learning methods and traditional shallow transfer learning methods. This validates that reducing the domain discrepancy by embedding domain-adaptation modules into deep networks (DDC, DAN, RevGrad, and RTN) can learn more transferable features. \textbf{(3)} The JAN models outperform previous methods by large margins and set new state of the art record. Different from all previous deep transfer learning methods that only adapt the marginal distributions based on independent feature layers (one layer for RevGrad and multilayer for DAN and RTN), JAN adapts the joint distributions of network activations in all domain-specific layers to fully correct the shifts in joint distributions across domains. Although both JAN and DAN \cite{cite:ICML15DAN} adapt multiple domain-specific layers, the improvement from DAN to JAN is crucial for the domain adaptation performance: JAN uses a JMMD penalty to reduce the shift in the joint distributions of multiple task-specific layers, which reflects the shift in the joint distributions of input features and output labels; DAN needs multiple MMD penalties, each independently reducing the shift in the marginal distribution of each layer, assuming feature layers and classifier layer are independent.

By going from AlexNet to extremely deep ResNet, we can attain a more in-depth understanding of feature transferability.
\textbf{(1)} ResNet-based methods outperform AlexNet-based methods by large margins. This validates that very deep convolutional networks, e.g. VGGnet \cite{cite:ICLR15VGG}, GoogLeNet \cite{cite:CVPR15GoogleNet}, and ResNet, not only learn better representations for general vision tasks but also learn more transferable representations for domain adaptation. 
\textbf{(2)} The JAN models significantly outperform ResNet-based methods, revealing that even very deep networks can only reduce, but not remove, the domain discrepancy.
\textbf{(3)} The boost of JAN over ResNet is more significant than the improvement of JAN over AlexNet. This implies that JAN can benefit from more transferable representations. 

The great aspect of JAN is that via the kernel trick there is no need to train a separate network to maximize the MMD criterion~\eqref{eqn:MMD} for the ball of a RKHS. However, this has the disadvantage that some kernels used in practice are unsuitable for capturing very complex distances in high dimensional spaces such as natural images \cite{cite:Arxiv17WGAN}.
The JAN-A model significantly outperforms the previous domain adversarial deep network \cite{cite:ICML15RevGrad}. The improvement from JAN to JAN-A also demonstrates the benefit of adversarial training for optimizing the JMMD in a richer function class. By maximizing the JMMD criterion with respect to a separate network, JAN-A can maximize the distinguishability of source and target distributions. Adapting domains against deep features where their distributions maximally differ, we can enhance the feature transferability.

\begin{figure*}[tbp]
  \centering
  \subfigure[DAN: \textit{Source}=\textbf{A}]{
    \includegraphics[width=0.17\textwidth]{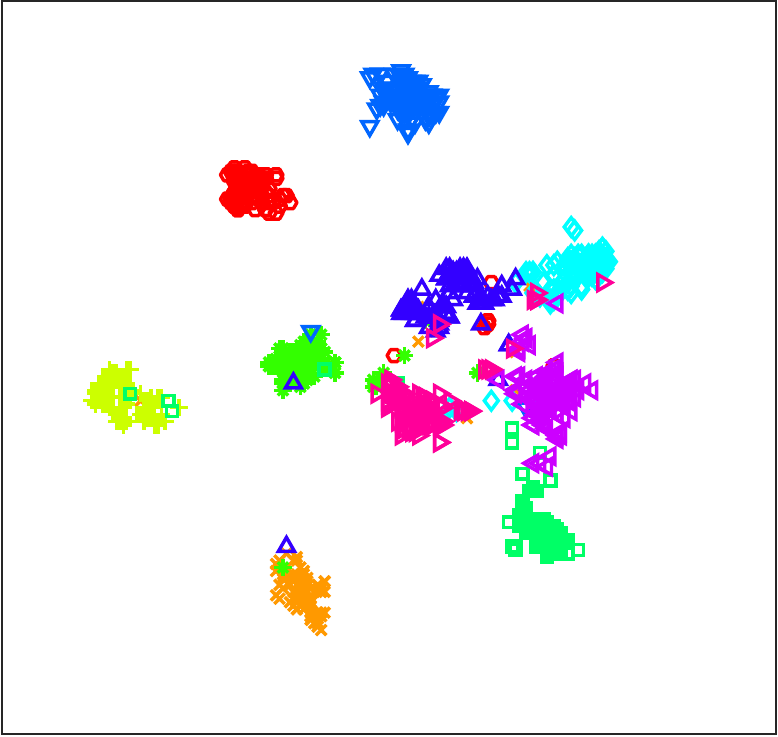}
    \label{fig:DANs}
  }\hfil
  \subfigure[DAN: \textit{Target}=\textbf{W}]{
    \includegraphics[width=0.17\textwidth]{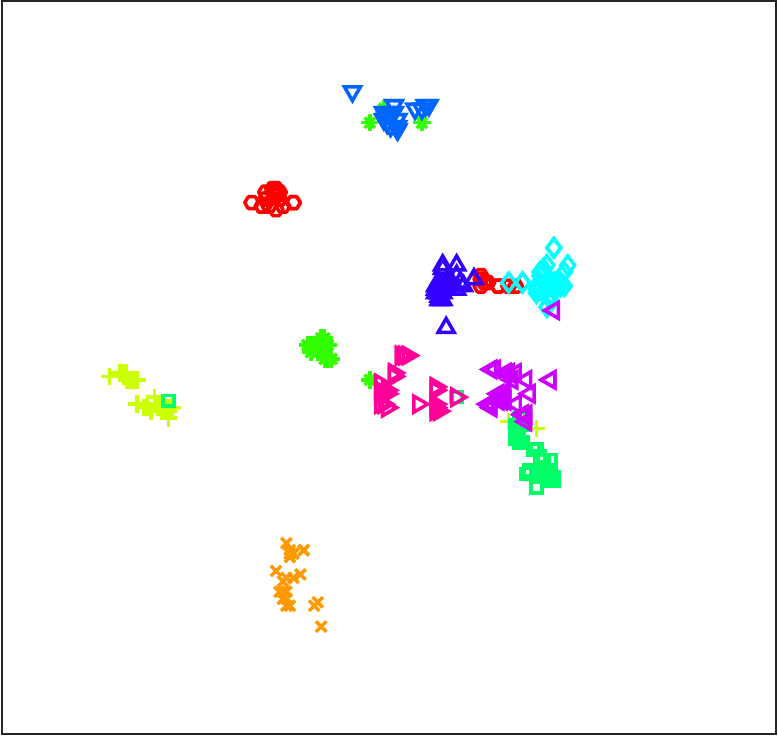}
    \label{fig:DANt}
  }\hfil
  \subfigure[JAN: \textit{Source}=\textbf{A}]{
    \includegraphics[width=0.17\textwidth]{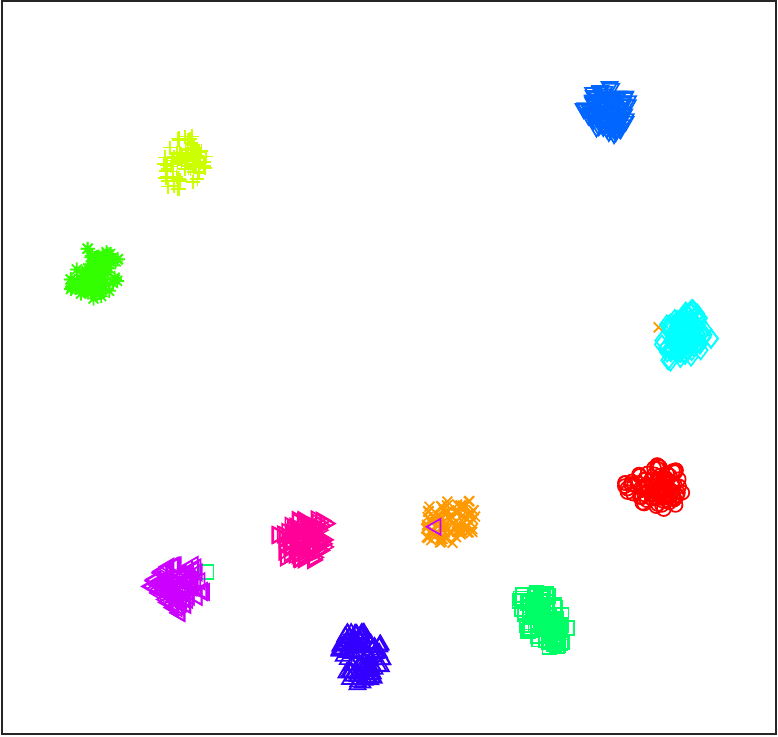}
    \label{fig:JANs}
  }\hfil
  \subfigure[JAN: \textit{Target}=\textbf{W}]{
    \includegraphics[width=0.17\textwidth]{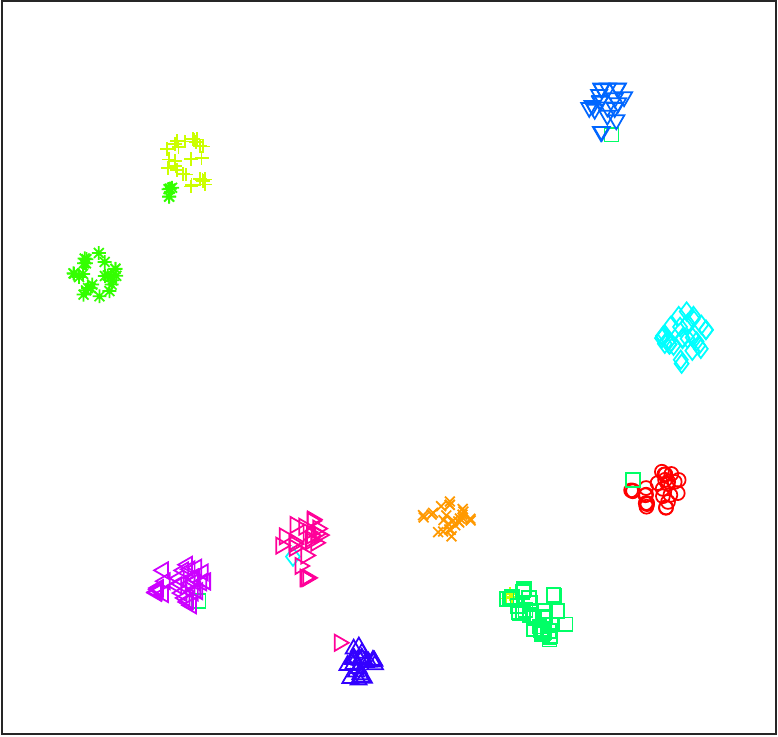}
    \label{fig:JANt}
  }
  \vspace{-10pt}
  \caption{The t-SNE visualization of network activations (ResNet) generated by DAN (a)(b) and JAN (c)(d), respectively.}
  \vspace{-10pt}
\end{figure*}

\begin{figure*}[tbp]
  \centering
  \subfigure[${\cal A}$-distance]{
    \includegraphics[width=0.21\textwidth]{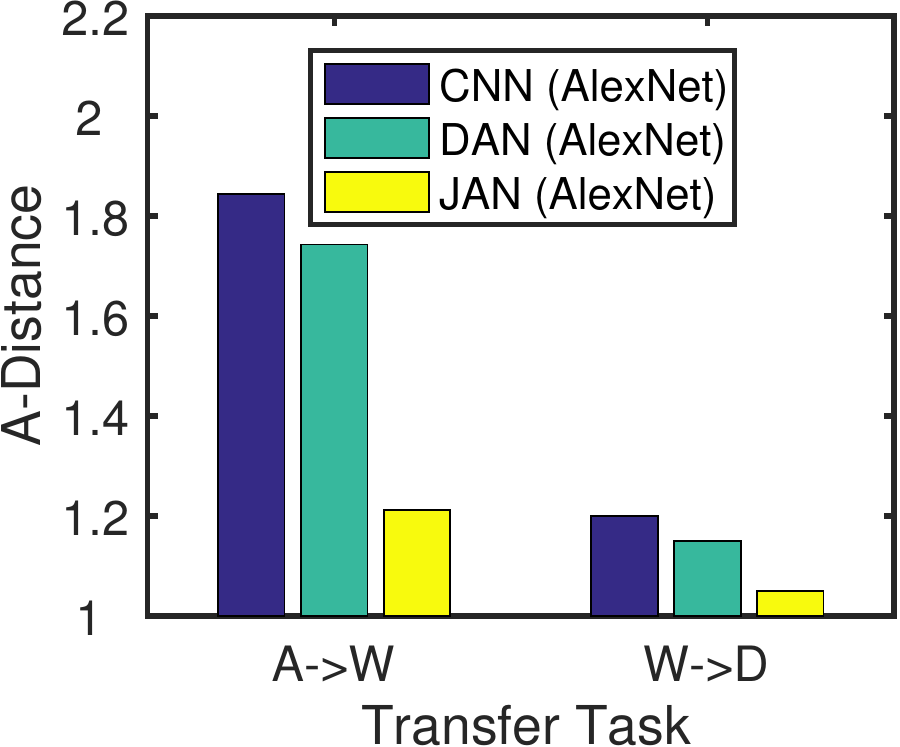}
    \label{fig:Adist}
  }
  \subfigure[JMMD]{
    \includegraphics[width=0.21\textwidth]{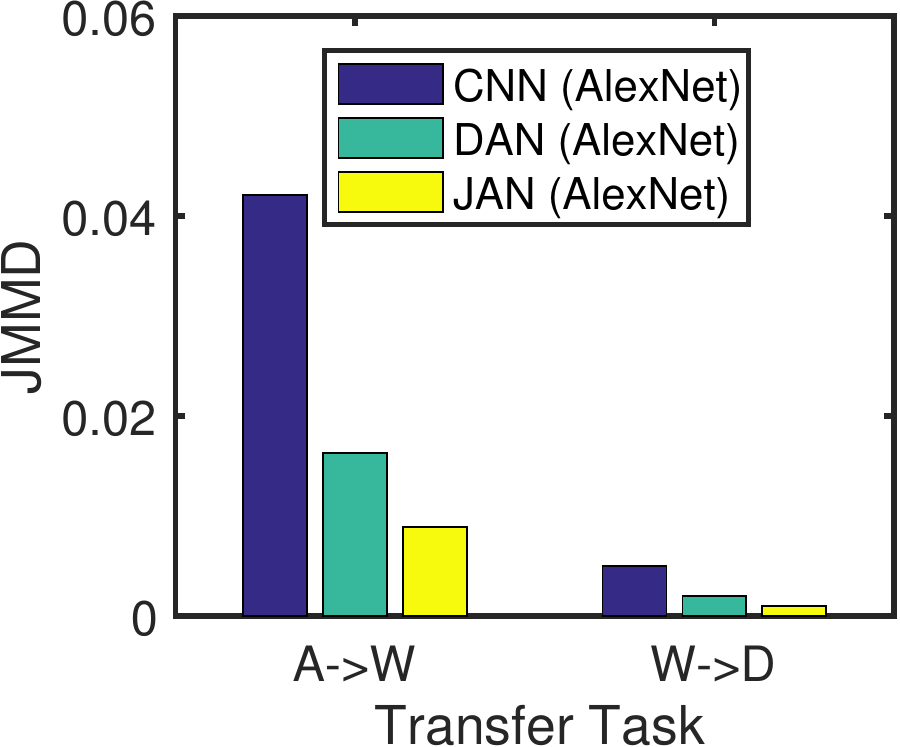}
    \label{fig:JMMD}
  }
  \subfigure[Accuracy w.r.t. $\lambda$]{
    \includegraphics[width=0.21\textwidth]{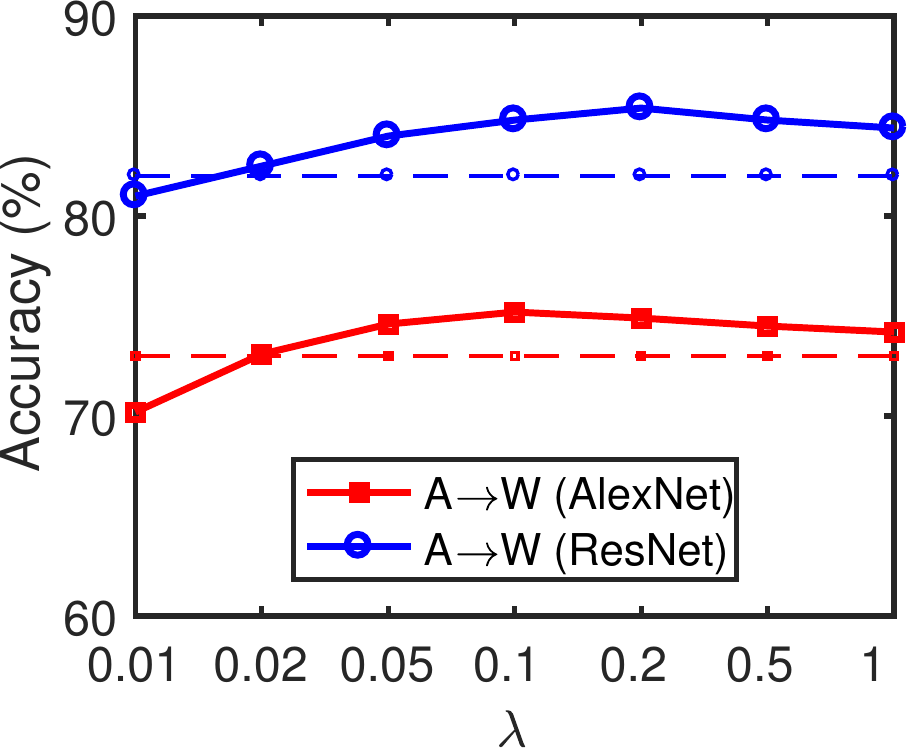}
    \label{fig:sensitivity}
  }
  \subfigure[Convergence]{
    \includegraphics[width=0.21\textwidth]{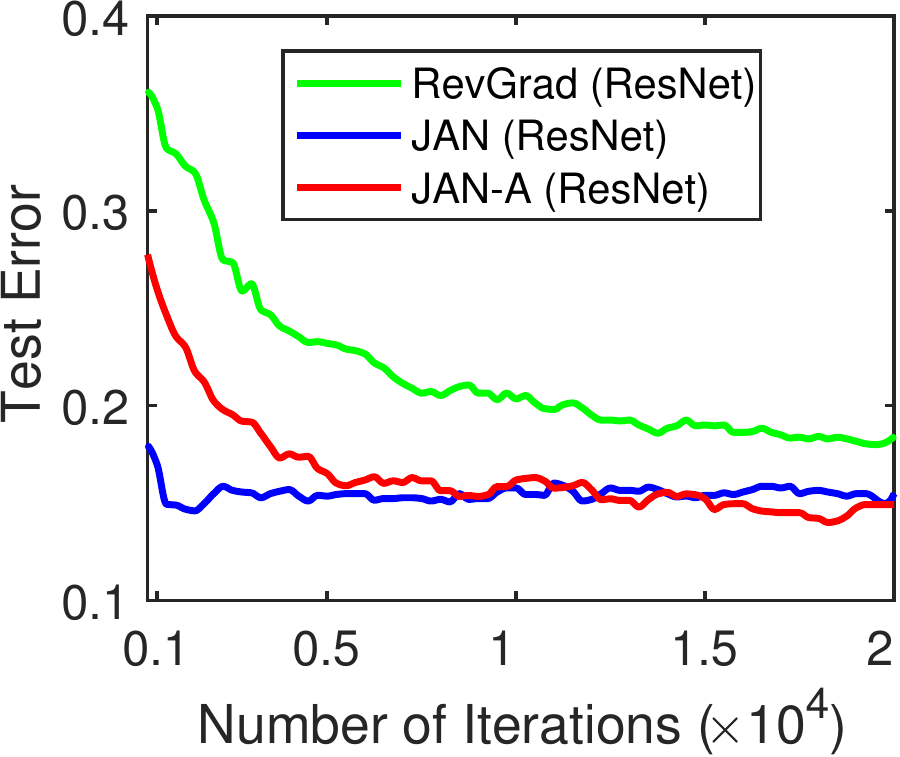}
    \label{fig:error}
  }
  \vspace{-10pt}
  \caption{Analysis: (a) ${\cal A}$-distance; (b) JMMD; (c) parameter sensitivity of $\lambda$; (d) convergence (dashed lines show best baseline results).}
  \vspace{-10pt}
\end{figure*}

The three domains in \textit{ImageCLEF-DA} are more balanced than those of \textit{Office-31}. With these more balanced transfer tasks, we are expecting to testify whether transfer learning improves when domain sizes do not change. The classification accuracy results based on both AlexNet and ResNet are shown in Table~\ref{table:imageclef-da}. The JAN models outperform comparison methods on most transfer tasks, but by less improvements. This means the difference in domain sizes may cause shift.

\subsection{Analysis}
\textbf{Feature Visualization:}
We visualize in Figures~\ref{fig:DANs}--\ref{fig:JANt} the network activations of task \textbf{A} $\rightarrow$ \textbf{W} learned by DAN and JAN respectively using t-SNE embeddings \cite{cite:ICML14DeCAF}. Compared with the activations given by DAN in Figure~\ref{fig:DANs}--\ref{fig:DANt}, the activations given by JAN in Figures~\ref{fig:JANs}--\ref{fig:JANt} show that the target categories are discriminated much more clearly by the JAN source classifier. This suggests that the adaptation of joint distributions of multilayer activations is a powerful approach to unsupervised domain adaptation.

\textbf{Distribution Discrepancy:}
The theory of domain adaptation~\cite{cite:ML10DAT,cite:COLT09DAT} suggests $\mathcal{A}$-distance as a measure of distribution discrepancy, which, together with the source risk, will bound the target risk. The proxy ${\cal{A}}$-distance is defined as ${d_{\cal A}} = 2\left( {1 - 2\epsilon } \right)$, where $\epsilon$ is the generalization error of a classifier (e.g. kernel SVM) trained on the binary problem of discriminating the source and target. Figure~\ref{fig:Adist} shows ${d_{\cal A}}$ on tasks \textbf{A} $\rightarrow$ \textbf{W}, \textbf{W} $\rightarrow$ \textbf{D} with features of CNN, DAN, and JAN. We observe that ${d_{\cal A}}$ using JAN features is much smaller than ${d_{\cal A}}$ using CNN and DAN features, which suggests that JAN features can close the cross-domain gap more effectively. As domains \textbf{W} and \textbf{D} are very similar, ${d_{\cal A}}$ of task \textbf{W} $\rightarrow$ \textbf{D} is much smaller than that of \textbf{A} $\rightarrow$ \textbf{W}, which explains better accuracy of \textbf{W} $\rightarrow$ \textbf{D}.

A limitation of the ${\cal A}$-distance is that it cannot measure the cross-domain discrepancy of joint distributions, which is addressed by the proposed JMMD \eqref{eqn:JMMDhat}. We compute JMMD \eqref{eqn:JMMDhat} across domains using CNN, DAN and JAN activations respectively, based on the features in $fc7$ and ground-truth labels in $fc8$ (the target labels are not used for model training). Figure~\ref{fig:JMMD} shows that JMMD using JAN activations is much smaller than JMMD using CNN and DAN activations, which validates that JANs successfully reduce the shifts in joint distributions to learn more transferable representations.

\textbf{Parameter Sensitivity:}
We check the sensitivity of JMMD parameter $\lambda$, i.e. the maximum value of the relative weight for JMMD. Figure~\ref{fig:sensitivity} demonstrates the transfer accuracy of JAN based on AlexNet and ResNet respectively, by varying $\lambda \in \{ 0.01, 0.02, 0.05, 0.1, 0.2, 0.5, 1 \}$ on task \textbf{A} $\rightarrow$ \textbf{W}. The accuracy of JAN first increases and then decreases as $\lambda$ varies and shows a bell-shaped curve. This confirms the motivation of deep learning and joint distribution adaptation, as a proper trade-off between them enhance transferability.

\textbf{Convergence Performance:}
As JAN and JAN-A involve adversarial training procedures, we testify their convergence performance. Figure~\ref{fig:error} demonstrates the test errors of different methods on task \textbf{A $\rightarrow$ W}, which suggests that JAN converges fastest due to nonparametric JMMD while JAN-A has similar convergence speed as RevGrad with significantly improved accuracy in the whole procedure of convergence.

\section{Conclusion}
This paper presented a novel approach to deep transfer learning, which enables end-to-end learning of transferable representations. Unlike previous methods that match the marginal distributions of features across domains, the proposed approach reduces the shift in joint distributions of the network activations of multiple task-specific layers, which approximates the shift in the joint distributions of input features and output labels. The discrepancy between  joint distributions can be computed by embedding the joint distributions in a tensor-product Hilbert space, which can be scaled linearly to large samples and be implemented in most deep networks. Experiments testified the efficacy of the proposed approach.

\section*{Acknowledgments}
We thank Zhangjie Cao for conducting part of experiments. This work was supported by NSFC (61502265, 61325008), National Key R\&D Program of China (2016YFB1000701, 2015BAF32B01), and Tsinghua TNList Lab Key Projects.

\bibliography{JAN}
\bibliographystyle{icml2017}

\end{document}